\pdfoutput=1

\documentclass[11pt]{article}

\usepackage{acl}

\usepackage{times}
\usepackage{latexsym}

\usepackage[T1]{fontenc}

\usepackage[utf8]{inputenc}

\usepackage{microtype}

\usepackage{inconsolata}

\usepackage{graphicx}
\usepackage{algorithm}
\usepackage{algorithmic}
\usepackage{xcolor}
\usepackage{arydshln}
\usepackage{amssymb}
\usepackage{amsmath}
\usepackage{booktabs}
\usepackage{dsfont}
\usepackage{multirow}
\usepackage{tikz}
\usepackage{kotex}

\usepackage{colortbl}
\usepackage{pifont}
\usepackage{tikz}
\usepackage{supertabular}
\usepackage{array}
\usepackage{tabularray}
\usepackage{ragged2e} %
\usepackage{enumitem}
\usepackage{longtable} %
\usepackage[most]{tcolorbox}

\title{Unveiling the Limits of Large Language Models in Inferring Pragmatic Meaning from Non-Verbal Responses}

\author{Sugyeong Eo$^{1}$, Heuiseok Lim$^{2}$\thanks{Corresponding author}\\
$^{1}$Department of Software, Yonsei University Mirae Campus, Republic of Korea\\
$^{2}$Department of Computer Science and Engineering, Korea University, Republic of Korea\\
\texttt{s.eo@yonsei.ac.kr \quad limhseok@korea.ac.kr}}

\begin{document}
\maketitle
\begin{abstract}
Although large language models (LLMs) have shown considerable progress in pragmatic language understanding, prior research has focused mainly on their comprehension of verbal behavior. Nonetheless, non-verbal behavior remains a fundamental component of human communication, especially when deliberately utilized in isolation to convey indirect meanings.
In this work, we present the first systematic evaluation of LLMs' ability to infer pragmatic meaning in dialogue consisting solely of non-verbal responses. We explore three research questions: (1) \textit{Can LLMs recognize indirect intent conveyed through non-verbal responses?} (2) \textit{When and how do LLMs fail to capture non-verbal intent?} (3) \textit{How can we improve LLMs' ability to interpret non-verbal intent?}. Through the evaluation, we observe that LLMs struggle to infer underlying meaning from non-verbal responses, with accuracy dropping by up to 60\% points compared to verbal ones. Further extensive analysis reveals a behavioral pattern in LLMs' interpretations of non-verbal behavior and demonstrates that in-context learning facilitates pragmatic inference.
\end{abstract}

\section{Introduction}
Recent years have demonstrated remarkable progress in large language models (LLMs), with strong generalization across diverse natural language processing tasks~\cite{NEURIPS2020_1457c0d6,ruis2023large,wei2022emergent,wei2022chain}. Building on this progress, an increasing number of studies have explored whether LLMs are capable of understanding non-literal language phenomena, such as implicature and indirect speech acts, which require sensitivity to cognitively grounded aspects of human communication \cite{grice1975logic,austin1975things,searle1975indirect,hu-etal-2023-fine, lee-etal-2025-pragmatic}. Notably, LLMs have demonstrated growing competence in pragmatic language understanding, extending their capabilities beyond surface-level understanding to the inference of implicit meaning~\cite{wu-etal-2024-rethinking, park-etal-2024-multiprageval}.

\begin{table}[]
\centering
\resizebox{\linewidth}{!}{%
\begin{tabular}{lcc|c}
\toprule[1.5pt]
 \textbf{Model} & \textbf{Verbal} & \textbf{Non-Verbal} & \textbf{$\Delta$} \\ \midrule[1.5pt]
3B-Llama & 0.96 & 0.37 & \cellcolor[HTML]{E06666}-0.59 \\
3.8B-Phi & 1.00 & 0.41 & \cellcolor[HTML]{E06666}-0.59 \\\cdashline{1-4}
8B-Llama & 0.88 & 0.37 & \cellcolor[HTML]{E47A7A}-0.51 \\
8B-Ministral & 0.48 & 0.46 & \cellcolor[HTML]{FDF9F9}-0.02 \\
7B-Qwen & 0.94 & 0.47 & \cellcolor[HTML]{E68585}-0.47 \\ 
12B-Mistral-NeMo & 0.96 & 0.38 & \cellcolor[HTML]{E06868}-0.58 \\
14B-Qwen & 0.78 & 0.31 & \cellcolor[HTML]{E68585}-0.47 \\
14B-Phi & 0.98 & 0.48 & \cellcolor[HTML]{E47D7D}-0.50 \\\cdashline{1-4}
32B-Qwen & 0.90 & 0.58 & \cellcolor[HTML]{EEACAC}-0.32 \\
72B-Qwen & 0.94 & 0.76 & \cellcolor[HTML]{F5D0D0}-0.18 \\
70B-Llama & 0.96 & 0.70 & \cellcolor[HTML]{F1BBBB}-0.26 \\\cdashline{1-4}
GPT-4.1-mini & 0.88 & 0.61 & \cellcolor[HTML]{F0B8B8}-0.27 \\
GPT-4o & 0.82 & 0.46 & \cellcolor[HTML]{ECA1A1}-0.36 \\
GPT-o3 & 0.96 & 0.73 & \cellcolor[HTML]{F2C2C2}-0.23\\
Claude-3.7-Sonnet & 0.94 & 0.54 & \cellcolor[HTML]{E99797}-0.40 \\ \cdashline{1-4}
Human & 0.99 & 0.91 & \cellcolor[HTML]{FBECEC}-0.08 \\ \bottomrule[1.5pt]
\end{tabular}%
}
\caption{Performance comparison of LLMs on verbal and non-verbal response settings. We report accuracy for both settings and their performance gap ($\Delta$).}
\label{tb:verbal}
\end{table}

Despite these advances, existing research has primarily focused on assessing LLMs' ability to interpret verbal responses, with non-verbal behavior typically considered only in conjunction with verbal input. However, in real-world contexts, non-verbal elements constitute a fundamental component of communication and often operate independently as exclusive signals of communicative intent~\cite{wharton2009pragmatics}. For instance, a person remaining silent after being asked ``Am I disturbing you?'' is pragmatically interpreted as an indirect indication that the speaker is causing a disturbance. Such non-verbal responses are deliberately employed to convey indirect intent.
While these interactions are intuitive to humans, it remains an open question whether LLMs are capable of recognizing communicative intent solely from non-verbal responses. 
Accurate understanding of intended non-verbal behaviors is especially important, as they are universally recognized communicative cues that play a vital role when spoken language is limited by linguistic boundaries~\cite{burgoon2021nonverbal}.

In this work, to the best of our knowledge, we present the first systematic evaluation of LLMs in understanding responses composed solely of non-verbal elements within dialogue. We categorize three types of non-verbal responses and assess model performance: silence, facial expressions, and movements. To guide our evaluation, we establish the following research questions: (1) \textit{Can LLMs recognize indirect intent conveyed through non-verbal responses?} (2) \textit{When and how do LLMs fail to capture non-verbal intent?} (3) \textit{How can we improve LLMs' ability to interpret non-verbal intent?}.

We conduct extensive experiments across six model families, covering parameter sizes from 3B to over 100 billion. Remarkably, although most LLMs achieve near-human performance in interpreting intent from verbal responses, their accuracy drops by as much as 60\% points when processing responses composed solely of non-verbal signals. Performance is notably poor in the silence category, where LLMs achieve only 0.45 accuracy compared to 0.91 for human-level performance, indicating a clear failure to capture implicit intent. Further analysis indicates that the majority of errors stem from abstract and literal descriptions, which restrict interpretations to surface-level features and fail to align with the underlying communicative intentions. Further investigation into the potential for enhancing this capability reveals that few-shot in-context learning substantially enhances the alignment of surface-level cues with the underlying communicative intentions. Our contributions are threefold:
\begin{itemize}
    \item To the best of our knowledge, this work provides the first systematic evaluation of LLMs' ability to infer underlying intent from dialogue responses composed exclusively of non-verbal signals.
    \item Comprehensive experiments spanning six prominent LLM families (3B–100B+) demonstrate that the models face persistent challenges in inferring underlying meaning, resulting in markedly diminished performance relative to their verbal counterparts.
    \item Further analysis reveals LLM behavioral patterns and shows that few-shot in-context learning enhances their ability to comprehend non-verbal intent.
\end{itemize}

\begin{table*}[]
\centering
\resizebox{\textwidth}{!}{%
\begin{tabular}{c|c|l}
\toprule[1.5pt]
\textbf{Catetory} & \multicolumn{1}{c}{\textbf{Options}} & \multicolumn{1}{|c}{\textbf{Content}} \\ \midrule
\multicolumn{2}{c|}{\textbf{\cellcolor[HTML]{e0dbef}Question}} & \cellcolor[HTML]{e0dbef}\begin{tabular}[c]{@{}l@{}}Choose the option that most appropriately interprets the context and underlying intent of \\ the conversation below. Assume that both parties faithfully engage in the conversation.\end{tabular} \\ \midrule
\multirow{6}{*}{\textbf{Silence}} 
&\cellcolor[HTML]{ced4da}Conversation & \cellcolor[HTML]{ced4da}A: Can you spare some time this weekend? I'm looking for someone to help me move. B: ...\\
 & \cellcolor[HTML]{FEE8A9}Option-1 & \cellcolor[HTML]{FEE8A9}\begin{tabular}[c]{@{}l@{}}B is indirectly rejecting A's request by deliberately remaining silent.\end{tabular} \\
 & Option-2 & A is asking B for help with moving, which is the act of relocating to a different place. \\
 & Option-3 & A is asking B for help with moving, but B has nothing to say and didn't respond. \\
 & Option-4 & In response to A's request, B is expressing a positive intention through silence. \\
 & Option-5 & A is asking a question, and B is not responding to it. \\\midrule
\multirow{6}{*}{\begin{tabular}[c]{@{}c@{}}\textbf{Facial}\\ \textbf{Expressions}\end{tabular}} &  
 \cellcolor[HTML]{ced4da}Conversation & \cellcolor[HTML]{ced4da}A: You know you can't miss our team dinner again, right? B: \includegraphics[height=1.1em]{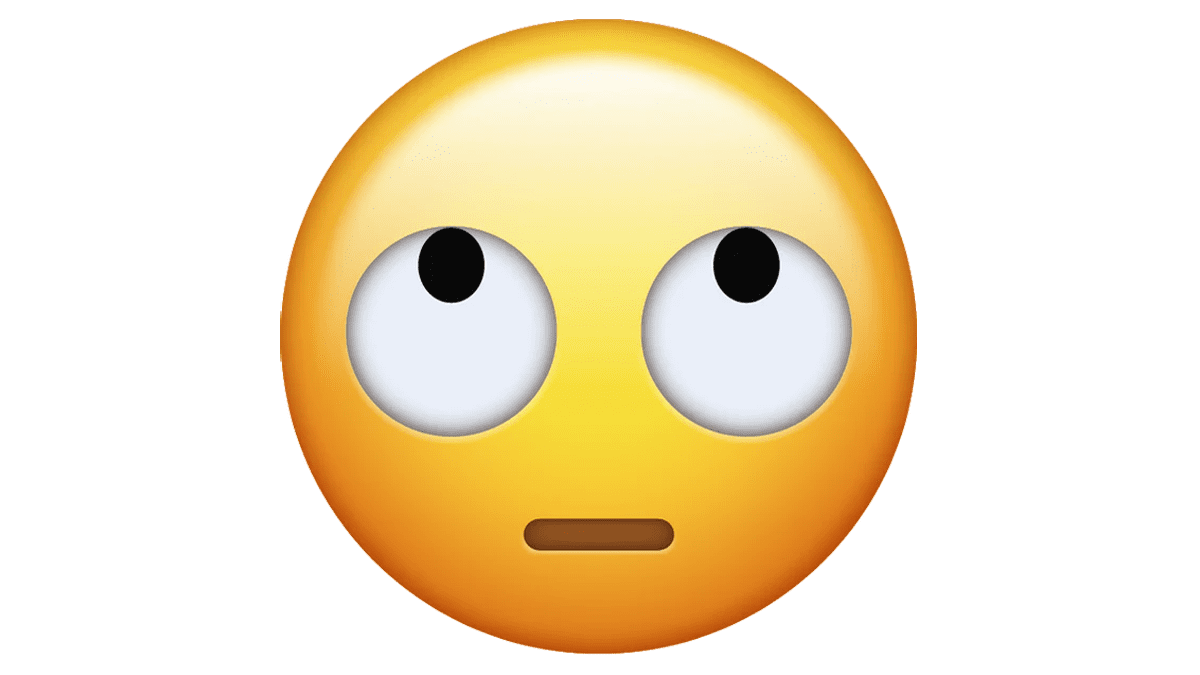}
\\
 & Option-1 & B looked up after A spoke. Maybe there was something on the ceiling. \\
 &\cellcolor[HTML]{FEE8A9} Option-2 & \cellcolor[HTML]{FEE8A9}B seems to be avoiding the team dinner through evasive actions. \\
 & Option-3 & B expresses the intention to participate in the gathering by raising their eyes after A spoke. \\
 & Option-4 & A is talking about team dinners, which help build team spirit. \\
 & Option-5 & A is asking a question, and B is showing only a facial expression in return. \\\midrule
\multirow{6}{*}{\textbf{Movements}}  &  \cellcolor[HTML]{ced4da}Conversation & \cellcolor[HTML]{ced4da}A: Starting today, write 10 pages of the report every day. B: (Points to the broken laptop.) \\
 & Option-1 & B agrees with A's statement but did not mention it directly, considering social propriety. \\
 & Option-2 & B, not fully understanding A's statement, pointed to a broken laptop to change the subject. \\
 & Option-3 & B picked up the broken laptop to fix it. \\
 & \cellcolor[HTML]{FEE8A9}Option-4 & \cellcolor[HTML]{FEE8A9}B indirectly expresses that they cannot do it because their laptop is broken, in response to A's statement. \\
 & Option-5 & A is asking a question, and B shows only an action in return. \\ \bottomrule[1.5pt]
\end{tabular}%
}
\caption{An example presenting the three evaluation categories of silence, facial expressions, and movements. The answer is highlighted in yellow.}
\label{tb:example}
\end{table*}

\section{Related Work}
\paragraph{Pragmatic Language Understanding}
Pragmatics is the systematic study of meaning that arises from language use, encompassing key theoretical constructs such as implicature, indirect speech acts, and deixis~\cite{grice1975logic,searle1975indirect,levinson1983pragmatics}. 
Recent efforts have investigated whether LLMs handle such phenomena by proposing new benchmarks~\cite{qi-etal-2023-pragmaticqa,sravanthi-etal-2024-pub,park-etal-2024-multiprageval}. 
Additionally, attempts have been made to enhance LLMs’ pragmatic abilities, including modifying training objectives~\cite{wu-etal-2024-rethinking} or designing prompts that target specific pragmatic phenomena~\cite{ruis2023the,lee-etal-2025-pragmatic}.
In parallel, researchers have analyzed common failures in LLMs’ pragmatic reasoning. For example, \citet{hu-etal-2023-fine} observe that LLMs often exhibit human-like error patterns, while \citet{sravanthi-etal-2024-pub} highlight particular difficulty with contrasting distractors.
Prior studies mainly focus on verbal cues, with non-verbal behavior typically serving as contextual support for interpreting verbal responses.

\paragraph{Understanding Non-verbal Cues in Language Models}
Non-verbal behavior refers to communicative signals that are conveyed without the use of words~\cite{knapp1972nonverbal,hall2019nonverbal}. Such signals include facial expressions, movements, prosody, and other behavioral cues that play a vital role in expressing intent during communication.
Given the communicative significance of non-verbal behavior, recent work has explored incorporating these signals into language models. For instance, \citet{lee2023developing} develop an empathetic LLM by conditioning it on non-verbal cues. Other approaches leverage emojis as affective indicators to infer emotional states from text~\cite{felbo-etal-2017-using}. 
\citet{hakami-etal-2023-arsarcasmoji} examine the role of facial expressions in detecting sarcasm. These studies highlight that non-verbal cues enrich pragmatic interpretation when combined with verbal input.
This study departs from previous approaches by focusing exclusively on scenarios where verbal utterances are entirely absent, and meaning must be inferred purely from non-verbal signals.

\section{Methodology}

\subsection{Problem Scope}
This study focuses on scenarios where responses consist solely of non-verbal behaviors. While paralinguistic cues such as pitch and amplitude are part of non-verbal communication, we exclude them from consideration as they mainly operate in conjunction with verbal behavior. To evaluate pragmatic competence, we instruct the model to \textit{``choose the option that most appropriately interprets the context and underlying intent of the conversation.''}. We further include the instruction \textit{``Assume that both participants are faithfully engaged in the conversation''} in the prompts, ensuring that non-verbal responses are intentionally structured to convey communicative intentions \citep{ekman1969nonverbal}. 
The evaluation examines how well LLMs infer intended meaning solely from non-verbal cues, without verbal utterances.

\subsection{Three Categories of Non-verbal Behavior}
Grounded in the theoretical foundations outlined in Appendix~\ref{app:category}, this study categorizes non-verbal behavior into facial expressions and bodily movements, while additionally conceptualizing silence as a distinct category owing to its absence of overt communicative signals.

\paragraph{Silence}
Silence is not merely the absence of speech, but rather constitutes a deliberate and strategic communicative act whose meaning must be inferred from context~\cite{johannesen1974functions,jaworski1992power,lane2002silence,bruneau1973communicative,jensen1973communicative}. This serves as a communicative signal, conveying implicit meanings such as refusal, deliberate avoidance, and tacit agreement. For instance, silence following a request often implies rejection, whereas silence in response to a question involving sensitive information typically indicates avoidance. Although it carries little explicit information, silence plays a critical role in conversational pragmatics as a powerful contextual signal.

\paragraph{Facial expressions}
Facial expressions serve as a visual symbolic form of non-verbal expression, often conveying emotions, intentions, or pragmatic nuances~\cite{dresner2010functions,hayati-etal-2019-sunny}. 
Despite their ubiquity in everyday communication, understanding the meaning of a facial expression requires contextual understanding.
For example, the facial expression `` \includegraphics[height=1.1em]{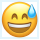} '' (smiling face with sweat) is often used to express embarrassment, or polite avoidance. Similarly, the facial expression ``\includegraphics[height=1.1em]{figure/rolling_eyes.png}'' (face with rolling eyes) typically conveys annoyance or disbelief, even though no verbal cue is provided. In this way, facial expressions, while visually simple, encode rich pragmatic meaning that must be inferred through contextual reasoning.

\paragraph{Movements}
Movements include gestures and postures, function as non-verbal cues that reflect a speaker’s psychological or interpersonal stance, playing a crucial role in conveying implicit intentions~\cite{patterson2012nonverbal,hall2019nonverbal,wharton2009pragmatics}. For instance, movements such as looking away may suggest intentional avoidance or discomfort, while scratching one's head often implies confusion, hesitation, or disagreement. The interpretation of such movements is highly context-dependent, as identical behaviors can convey different pragmatic meanings depending on situations.

\subsection{Evaluation Protocol}
We evaluate LLMs’ ability to interpret non-verbal responses using a set of multiple-choice questions, each consisting of five answer options. Given a prompt comprising a \texttt{[question]}, \texttt{[a dialogue context with a non-verbal response]}, and \texttt{[five candidate options]}, the model is tasked with selecting the interpretation that most accurately reflects the conversational context and the underlying communicative intent.

Distractors are carefully designed to represent four incorrect types: (1) Misinterpretations, which involve misunderstanding the intended meaning of the response~\cite{sravanthi-etal-2024-pub}; (2) Literal explanations, which rely on literal readings while disregarding the pragmatic context~\cite{hu-etal-2023-fine}; (3) Lexically coherent but semantically divergent responses, which display surface-level lexical similarity but lack contextual relevance~\cite{shisen-etal-2024-large,zheng-etal-2021-grice}; and (4) Abstract explanations, which consist of vague descriptions that even fail to address the core topic of the dialogue~\cite{peters2025generalization}.

The representation scheme for non-verbal responses is formulated on the basis of two principal considerations. To enable rigorous evaluation of an LLM's understanding of non-verbal behavior, we clearly separate the form of non-verbal responses from verbal ones and define an independent response format. At the same time, in order to encompass both general conversational contexts and web chat environments, response types are specified to accommodate both forms of communication. We represent responses in the silence category as `...’~\cite{wu2025yet, li-etal-2017-dailydialog}, facial expressions using Unicode emojis~\cite{felbo-etal-2017-using, hakami-etal-2023-arsarcasmoji, hu2017spice}, and movements as descriptive actions enclosed in parentheses~\cite{dirik2021cuegen, chen-etal-2022-summscreen}.
Descriptions of the detailed dataset construction, quality control, and statistics are provided in Appendix~\ref{app:data_construction}, while an illustrative example is presented in Table~\ref{tb:example}.

\begin{figure*}
\centering
\includegraphics[width=\linewidth]{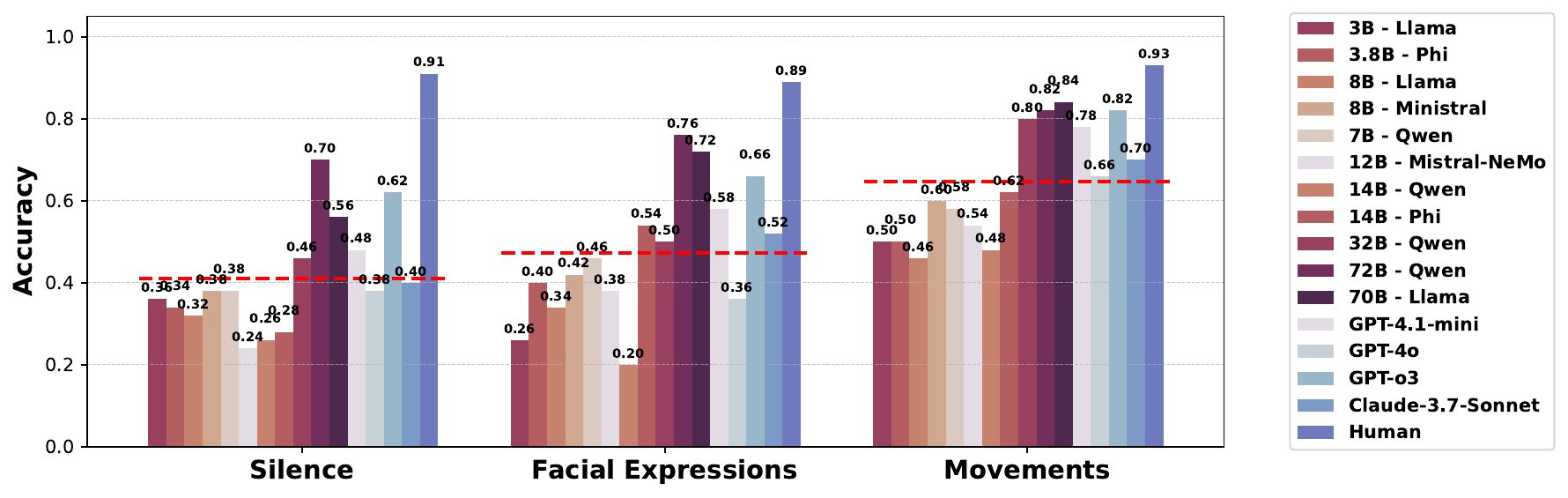}
\caption{Evaluation of LLM performance across three non-verbal response categories: silence, facial expressions, and movements. The red dotted line represents the average performance of all models within the category.} \label{fig:main}     
\end{figure*}

\section{Experiments}
This section reports a systematic evaluation of LLMs' ability to interpret non-verbal behavior, structured around three guiding research questions:
(1) Can LLMs recognize indirect intent conveyed through non-verbal responses? (2) When and how do LLMs fail to capture non-verbal intent? (3) How can we improve LLMs' ability to interpret non-verbal intent?

\subsection{Evaluation Setup}
We employ publicly available instruction-tuned language models to assess their ability to infer implicit intentions conveyed through non-verbal cues. Each input is framed as a multiple-choice question comprising five candidate options, and the model is tasked with selecting the most appropriate option. 
To better reflect real-world scenarios, all experiments are conducted in a zero-shot setting.

We conduct comprehensive experiments across 13 models with six different model families. The HuggingFace checkpoints used in this study are provided as follows:
\begin{itemize}[itemsep=1pt, topsep=2pt, parsep=0pt, partopsep=0pt]
\setlength\itemsep{1pt}
\setlength\parskip{0pt}
\setlength\parsep{0pt}
\item \textbf{Qwen}: 7B (\texttt{Qwen/Qwen2.5-7B-Instruct}), 14B (\texttt{Qwen/Qwen2.5-14B-Instruct}), 32B (\texttt{Qwen/Qwen2.5-32B-Instruct}), 72B (\texttt{Qwen /Qwen2.5-72B-Instruct})
\item \textbf{Mistral}: 8B (Ministral) (\texttt{mistralai/Mini stral-8B-Instruct-2410}), 12B (NeMo) (\texttt{mistralai/Mistral-Nemo-Instruct-2407})
\item \textbf{Llama}: 3B (\texttt{meta-llama/Llama-3.2-3B- Instruct}), 8B (\texttt{meta-llama/Llama-3.1- 8B-Instruct}), 70B (\texttt{meta-llama/Llama- 3.3-70B-Instruct})
\item \textbf{Phi}: 3.8B (\texttt{microsoft/Phi-3-mini-4k- instruct}), 14B (\texttt{microsoft/phi-4})
\item \textbf{API-based models}: \texttt{GPT-4.1-mini}, \texttt{GPT-4o}, \texttt{GPT-o3}, \texttt{Claude-3.7-Sonnet}
\end{itemize}

To ensure consistency across experiments, we uniformly apply a decoding temperature of 0.0 for deterministic outputs and limit the maximum input length to 512 tokens for all models. All evaluations are conducted on four NVIDIA A6000 GPUs (48GB). With regard to the human evaluation setup for comparison with model scores, six undergraduate students are recruited and asked to complete the tasks under identical experimental conditions\footnote{Annotators were compensated at rates exceeding the legally mandated minimum wage in the respective countries.}. Since the dataset encompasses everyday social interactions, no restrictions are placed on the annotators' areas of expertise. In addition, as the dataset includes globally recognized communicative practices, annotators are recruited from diverse national backgrounds (American, Korean, Vietnamese, and Bangladeshi), all of whom are proficient in English. The inter-annotator agreement is 0.7924 as measured by Krippendorff’s alpha, demonstrating substantial agreement among the annotators.

\subsection{RQ1. Can LLMs recognize indirect intent conveyed through non-verbal responses?}

\paragraph{Performance of LLMs across categories}
Figure~\ref{fig:main} reports the performance of language models across three categories of non-verbal responses. Among the three categories, the movement category shows the highest average accuracy of 0.67, followed by facial expressions with 0.51 and silence with 0.45. While the performance on movement-based cues appears relatively promising, it remains substantially below the human-level accuracy of 0.93, highlighting the considerable gap between model and human comprehension of implicit intent. The notably low performance in the silence condition is attributed to communicative richness, which poses greater challenges for inference by LLMs. In contrast to movements and facial expressions, the silence category conveys little to no explicit information, substantially increasing the difficulty of pragmatic inference for language models.
The findings indicate that, while humans are adept at interpreting subtle social cues, LLMs face difficulties, especially when the response provides minimal explicit information.

Notably, even the widely used GPT-4o demonstrates subpar performance in these settings. It achieves accuracies of 0.38, 0.36, and 0.66 in the silence, facial expressions, and movements, respectively, which is comparable to the performance observed in open-source 14B models. These suggest that, despite their advanced reasoning and factual competence, even highly capable LLMs struggle to interpret cues composed solely of non-verbal behavior. Even the thinking model GPT-o3 achieves performance comparable to that of the Llama 70B model, yet both fall significantly short of human-level accuracy, highlighting the challenges.

\begin{figure}
\centering
\includegraphics[width=\linewidth]{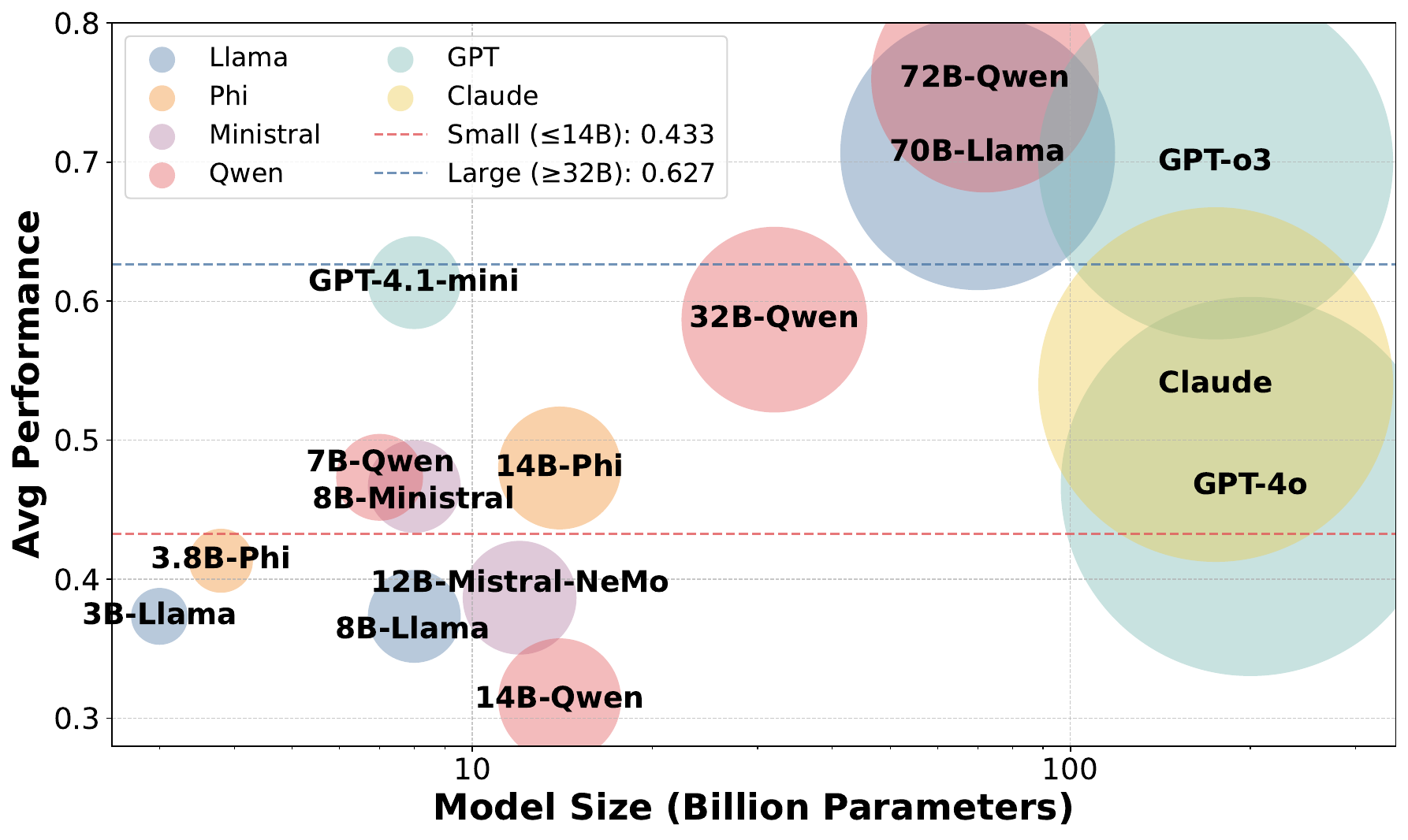}
\caption{Distribution of average performance across LLMs of varying parameter scales} \label{fig:average}     
\end{figure}

\begin{figure*}
\centering
\includegraphics[width=.95\linewidth]{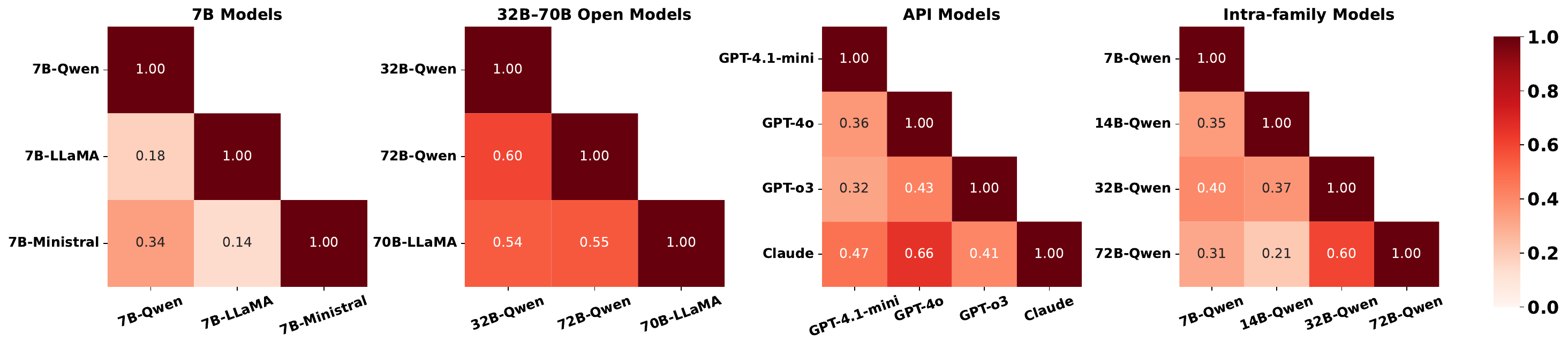}
\caption{Heatmap of prediction correlations across language models. The heatmap visualizes response-level agreement between models, grouped by model family and parameter size.} \label{fig:heatmap}     
\end{figure*}

\paragraph{The effect of parameter scale on model performance}
Figure~\ref{fig:average} visualizes the average performance of each model across the three non-verbal categories. The results show that models in the 3B to 14B range exhibit relatively lower performance, whereas larger models with 32B and 70B parameters tend to achieve higher accuracy. In particular, models from the Llama, Qwen, and Phi families display a consistent trend of improved performance as parameter size increases. These findings imply that scaling up parameter size enhances a model’s capacity to infer intentions from non-verbal responses. However, this pattern does not extend to API-based models. GPT-4.1-mini surpasses GPT-4o, and Claude and GPT-4o underperform relative to their scale. These observations suggest that, while increasing model size within the same architecture generally leads to improved performance, additional external factors may also play a non-negligible role in shaping model effectiveness.

\paragraph{Correlational patterns in model predictions}
To better understand the behavioral similarity among language models, we analyze prediction correlations based on response-level outputs, grouped by model family and parameter scale. The results depicted in Figure~\ref{fig:heatmap} show that 7B-scale models exhibit less consistent response patterns, suggesting that they lack sufficiently consolidated knowledge for interpreting non-verbal behavior.
Models with 32B to 70B parameters tend to exhibit increased positive correlations, suggesting a degree of shared underlying reasoning strategies. However, this also implies that such models show similar error patterns, pointing to potential shared limitations in their pragmatic inference capabilities.

\begin{figure}
\centering
\includegraphics[width=\linewidth]{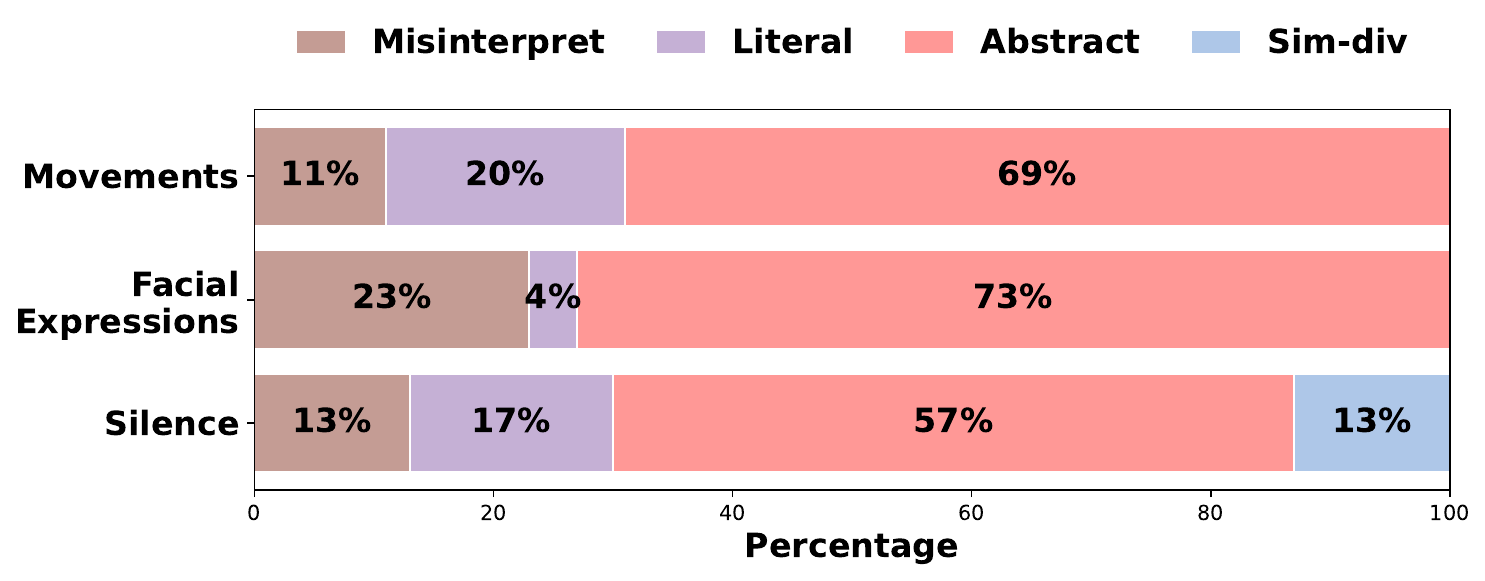}
\caption{Distribution of errors across four distractor types: misinterpretation, literal explanation, abstract explanation, and lexically coherent but semantically divergent response (sim-div). The analysis is based on the results of Llama-70B, Qwen-32B, and GPT-4o.} \label{fig:err_ana}     
\end{figure}

We observe that intra-family prediction correlations tend to increase with parameter size, further supporting the finding that larger models exhibit more aligned interpretive behavior when processing non-verbal signals. In conclusion, the predictive tendencies are more strongly influenced by parameter scale than by shared architectural similarity. 

\paragraph{Evaluating model performance in verbal and non-verbal response settings}\label{para:verbal}
The ability of LLMs to infer contextual intent from non-verbal-only responses remains limited, particularly in categories such as silence and facial expressions. These findings raise the possibility that the difficulty does not lie solely in the type of communication, but rather in a broader limitation of LLMs to capture indirect and implicit contextual cues.
To further explore this hypothesis, we conduct an additional experiment using a verbal setting, in which the responses are indirect but explicitly verbalized. An example is: ``\texttt{A: Can you spare some time this weekend? I'm looking for someone to help me move. B: Oh, my legs hurt,}'' where speaker B indirectly declines the request by offering an excuse rather than a direct refusal.

As shown in Table~\ref{tb:verbal}, LLMs demonstrate notably strong performance under the verbal condition. Even relatively small models, such as those with 3B or 8B parameters, achieve near-human performance, with the exception of the 8B-Ministral. These results indicate that LLMs correctly infer intent when conveyed through explicit verbal responses. In contrast, performance in the non-verbal condition shows a significant decrease. Models with 3B parameters exhibit performance drops of up to 0.59, and even models with 32B to 70B parameters report a reduction of up to 0.32. This degradation is also observed in API-based models. In comparison, human performance remains nearly consistent across both settings, suggesting that current LLMs still have room for improvement.

\begin{figure*}
\centering
\includegraphics[width=\linewidth]{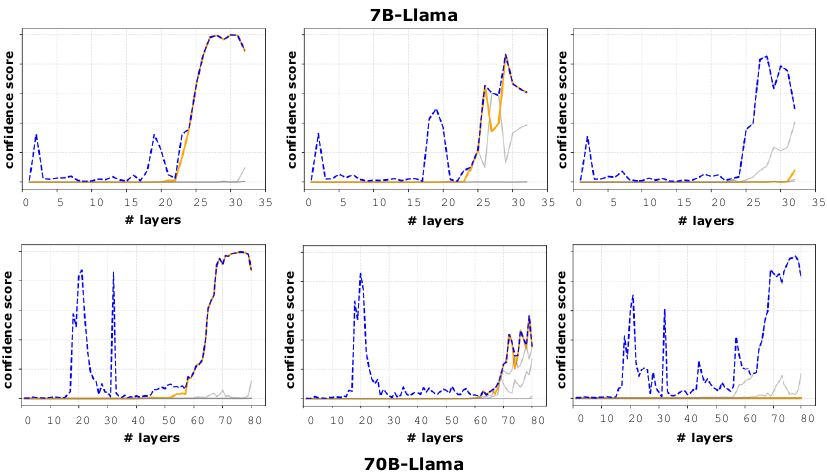}
\caption{Layer-wise probability analysis via logit lens. The blue dashed line shows the top-1 probability across layers, with the correct option in orange and the distractor options in gray.} \label{fig:logit}
\end{figure*}

\subsection{RQ2. When and how do LLMs fail to capture non-verbal intent?}

\paragraph{Error analysis}
Our focus shifts to a deeper exploration of the error patterns that emerge in LLMs’ interpretation of non-verbal responses. We analyze the selection behavior of LLMs across four predefined distractor types, and the corresponding error distributions are presented in Figure~\ref{fig:err_ana}. The results reveal that abstract explanations constitute the predominant source of error, accounting for more than half of the cases across all categories. 
Even when explicitly instructed to select the option that best reflects the dialogue context and underlying communicative intent, the models often default to narrowly oversimplified descriptions. This pattern reflects a tendency of the models to favor broadly applicable yet minimally variable responses~\cite{peters2025generalization}, limiting their ability to engage in deeper inferential reasoning. A related error type is literal interpretation, in which the responses become granular but still fail to capture the intended meaning. Taken together, these findings suggest that the primary source of the models' limited pragmatic competence lies in their focus on shallow and surface-level representations over the recognition and reasoning of latent intentions embedded in dialogue. In contrast, as shown in Table~\ref{tb:verbal}, the models exhibit comparatively strong performance on verbal responses, suggesting that the core issue is not an inherent deficit in pragmatic reasoning itself but rather insufficient grounding for non-verbal signals. Meanwhile, misinterpretation accounts for approximately 11\% to 23\% of all errors and is predominantly elicited by contrastive distractors, which is consistent with previous findings reported by \citet{sravanthi-etal-2024-pub}. 

\begin{figure}
\centering
\includegraphics[width=\linewidth]{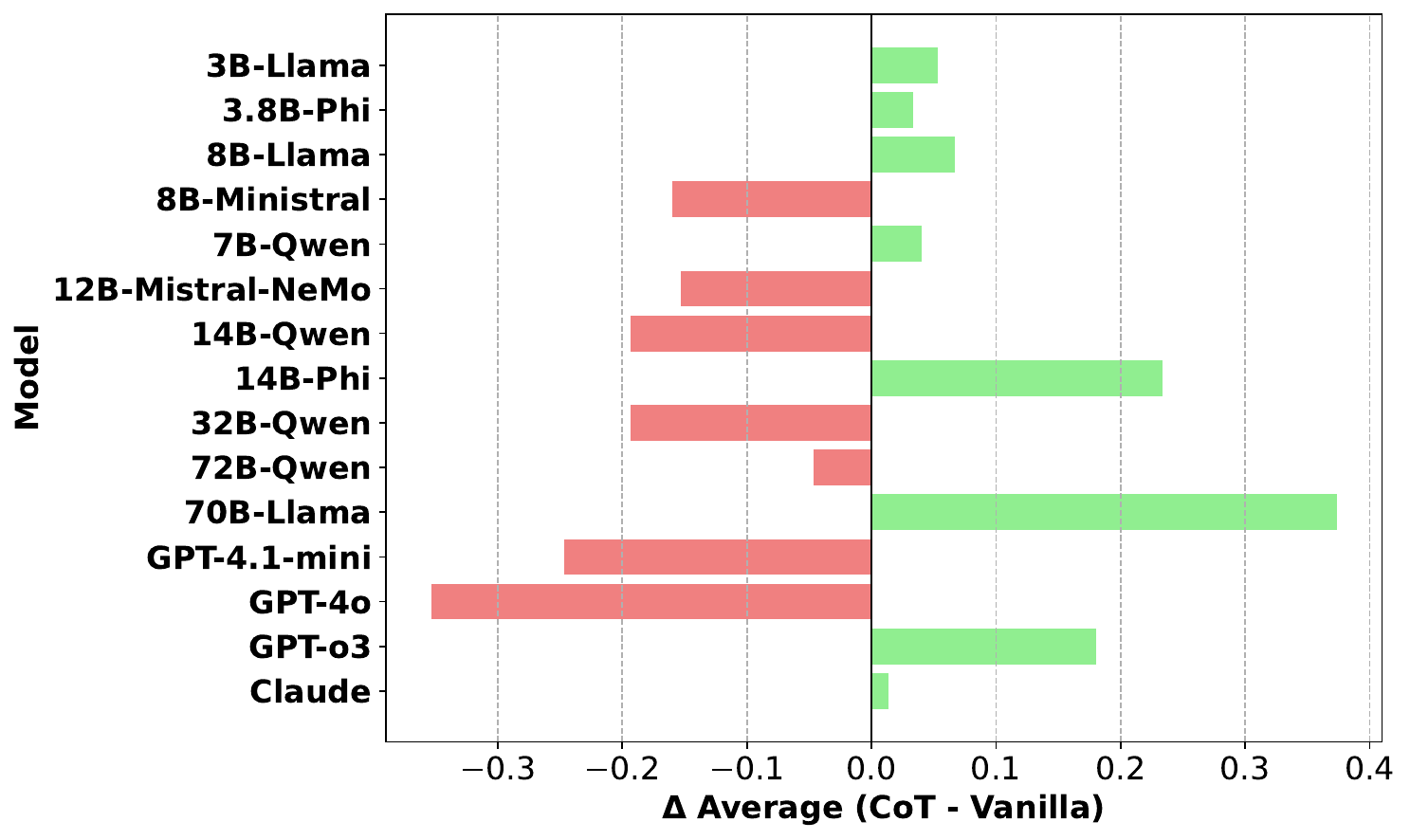}
\caption{Average performance change after applying CoT prompting, computed by subtracting each model's vanilla score from its corresponding CoT score} \label{fig:cot}     
\end{figure}

\begin{figure*}
\centering
\includegraphics[width=\linewidth]{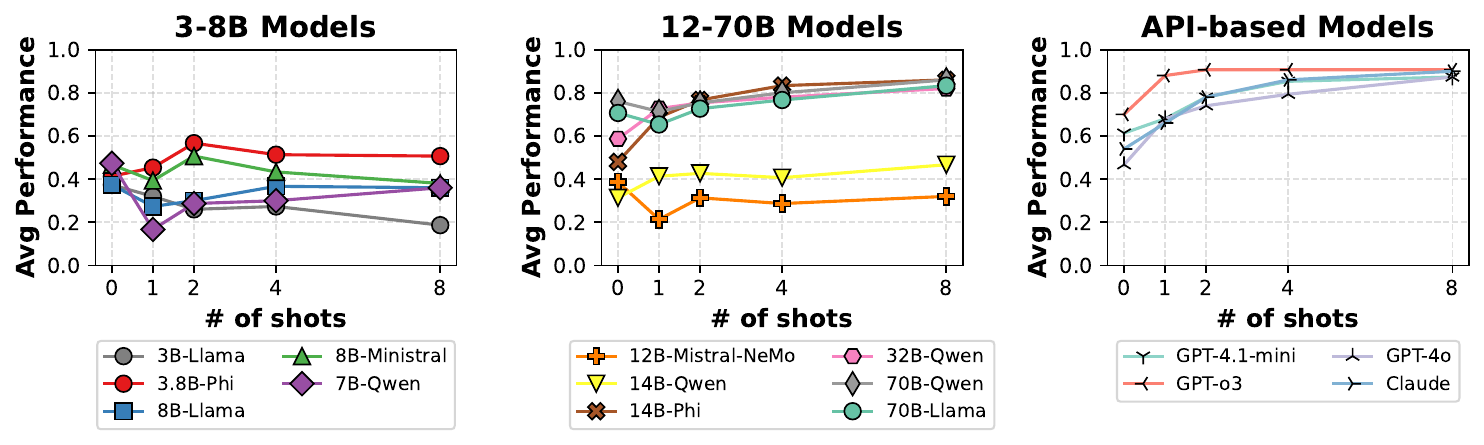}\caption{Average model accuracy across varying numbers of few-shot examples. Models are reported into three groups based on parameter size and access type.} \label{fig:fewshot}     
\end{figure*}

\begin{figure}
\centering
\includegraphics[width=\linewidth]{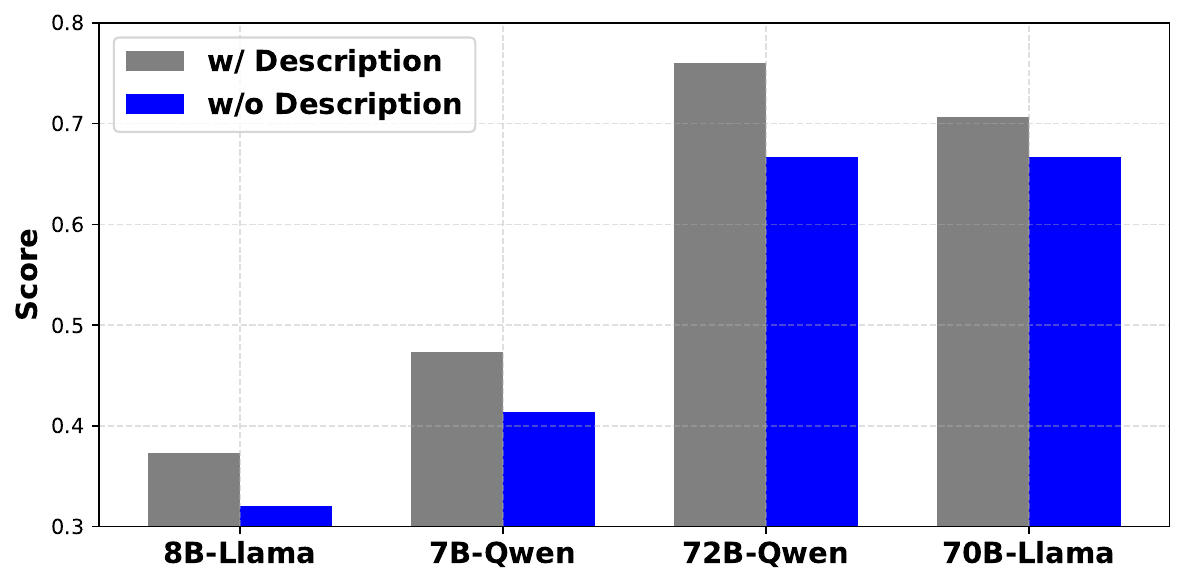}
\caption{Comparison of model performance with vs. without context descriptions} \label{fig:real_world} 
\end{figure}

\paragraph{Behavioral analysis of the model}
To advance a deeper analytical understanding, we investigate the internal behavioral patterns of LLMs. We aim to examine from which layer the model tends to attend to pragmatic cues, and how internal patterns emerge during the prediction process. To this end, we adopt the logit lens~\cite{nostalgebraist2020logitlens,halawi2024overthinking}, which facilitates layer-wise analysis of prediction dynamics within the model. We compute the token probability distribution at each layer $l$ by extracting the hidden state corresponding to the final token position, denoted as $\textbf{h}_{-1}^{(l)}\in\mathbb{R}^d$. After applying layer normalization and the language modeling head, the resulting distribution is obtained as: $\mathbf{p}^{(l)} = \text{Softmax} \left( \mathbf{W}_{\text{LM}} \cdot \text{LayerNorm}(\mathbf{h}^{(l)}_{-1}) + \mathbf{b}_{\text{LM}} \right)$.
For a fixed candidate set \( T = \{t_1, ..., t_5\} \) representing the tokenized options 1 to 5, we extract the confidence assigned to each candidate token $P^{(l)}_t = \mathbf{p}^{(l)}[t]$.
We further extract top-1 prediction at each layer, defined as the token with the highest probability and its corresponding confidence score $t^{(l)}_{\text{top1}} = \arg\max_{t} \mathbf{p}^{(l)}[t], \quad
P^{(l)}_{\text{top1}} = \max_{t} \mathbf{p}^{(l)}[t]$.

As illustrated in Figure~\ref{fig:logit}, confidence scores for the five candidate options remain uniformly low in the early layers and begin to rise significantly in the middle to later layers. This trend suggests that the model’s ability to interpret and infer context is primarily established in the deeper layers, where fine-grained semantic representations are formed. 

The top two subfigures illustrate correct predictions where the answer choice aligns with the top-1 probability. The model assigns sharply higher confidence to the correct option while effectively marginalizing alternative choices, indicating a clear resolution of the decision. In contrast, the bottom two subfigures illustrate incorrect predictions, revealing distinct error patterns. The model confidently assigns high probability to the wrong option, reflecting a complete misinterpretation of pragmatic cues. In the middle two figures, the model distributes low confidence across multiple candidates. Compared to high-confidence predictions, these examples exhibit delayed convergence and heightened instability in confidence patterns, particularly in the deeper layers. The results indicate that, much like human respondents, the model exhibits patterns of hesitation, confident correctness, and misinterpretation that depend on its reading of pragmatic cues.

\subsection{RQ3. How can we improve LLMs' ability to interpret non-verbal intent?}

\paragraph{Chain-of-Thoughts prompting}
This study investigates the potential to enhance the model's capacity for interpreting non-verbal intent. To this end, we apply Chain-of-Thought (CoT) prompting strategy to facilitate step-by-step reasoning. 
As shown in Figure~\ref{fig:cot}, most models demonstrate a trend in which the magnitude of performance degradation exceeded that of improvement, with GPT-4o notably declining by more than 0.3. We find that GPT-4o frequently favors literal interpretations, suggesting a limited ability to understand contextual cues even under CoT prompting.
In contrast, Llama-70B shows the most significant improvement, with consistent gains observed across most Llama and Phi models. Mistral and Qwen families generally show performance degradation, indicating that the effect of CoT varies by model family.

\paragraph{In-context learning}
We extend our investigation to few-shot prompting, a well-established approach for enhancing in-context learning. As shown in Figure~\ref{fig:fewshot}, in-context learning leads to notable performance gains, particularly among API-based models. GPT-o3, in particular, shows substantial improvement, suggesting that the model internalizes relational patterns from in-context examples well.
The 70B model also showed progressively enhanced performance as the number of shots increased. Smaller models (3B to 7B) exhibit slight performance declines, while the 14B model shows only marginal improvement. These findings indicate that larger models possess the representational capacity necessary to benefit from few-shot prompting, facilitating a more proper interpretation of non-verbal responses.

\subsection{Evaluating model performance without context description}
Our original prompt includes an explicit instruction ``\texttt{Assume that both parties faithfully engage in the conversation.}'', guiding the model to interpret non-verbal responses as intentional rather than incidental.
To approximate real-world conditions, we ablate this contextual description and compare the results, as shown in Figure~\ref{fig:real_world}. All four evaluated models, spanning two families and parameter scales, show decreased performance. These findings suggest that, in real-world use cases where the models are required to operate solely based on the given input without access to any additional context or few-shot examples, inferring intent from non-verbal content becomes substantially more challenging. 

\section{Conclusion}
Interpreting indirect meaning embedded in non-verbal behavior constitutes a critical step in advancing the pragmatic competence of LLMs. This study provided the first systematic evaluation of LLMs in interpreting communicative intent conveyed exclusively through non-verbal responses. Experimental results indicated a significant decline in performance under non-verbal conditions relative to verbal settings. A detailed analysis further revealed recurrent error types and behavioral patterns. Moreover, few-shot in-context learning was shown to enhance performance, indicating that LLMs acquire pragmatic mappings when supplemented with additional examples. By introducing a new dimension of pragmatic reasoning, this work extends the scope of LLM evaluation to encompass non-verbal communication.

\section*{Limitations}
This study is subject to several limitations. First, the evaluation is conducted exclusively in English and relies on pragmatic inferences that are generally assumed to hold across cultures. We do not examine how models interpret non-verbal behavior across different languages. Extending this work to a multilingual setup would offer valuable insights into language-specific model behavior and generalization.

Second, non-verbal behaviors can be perceived through various modalities.  We focus solely on textual input, since the language model is ultimately responsible for interpreting the speaker’s intent and context, regardless of the modality through which non-verbal signals are received. Nevertheless, incorporating multimodal inputs in future work will enhance the applicability of LLMs to real-world human–AI interactions.

\section*{Acknowledgments}
This research was supported by Basic Science Research Program through the National Research Foundation of Korea(NRF) funded by the Ministry of Education(NRF-2021R1A6A1A03045425) and this work was supported by the Commercialization Promotion Agency for R\&D Outcomes(COMPA) grant funded by the Korea government(MSIT)(2710086166) and this work was supported by Institute for Information \& communications Technology Promotion(IITP) grant funded by MSIT (RS-2024-00398115, Research on the reliability and coherence of outcomes produced by Generative AI) and this work was supported by IITP-ICT Creative Consilience Program grant funded by MSIT(IITP-2026-RS-2020-II201819).

\appendix

\bibliography{custom}

\appendix
\section{Categorizing Non-verbal Behavior}\label{app:category}
Non-verbal communication involves the transmission of messages through behavioral signals~\cite{mccornack2022choices}. This transmits meaning even when people encounter language barriers, making their comprehension within conversational contexts crucial~\cite{burgoon2021nonverbal}.

Non-verbal behavior has been categorized into distinct types in prior studies. 
For example, \citet{birdwhistell1952introduction} and \cite{DANESI2006207} define kinesics as the study of non-verbal behavior and categorize it into gestures, facial expressions, eye behavior, touch, and posture.
\citet{carmichael2023connecting} delineate four primary channels: touch, vocal tone, facial expressions, and bodily gestures. \citet{urakami2023nonverbal} classify human sensory channels by identifying body movements, gestures, and facial cues as components of the visual channel, and vocalizations and sounds as belonging to the auditory channel. \citet{wharton2009pragmatics} propose a more streamlined taxonomy, reducing non-linguistic cues to facial expressions and gestures.

Building upon this theoretical foundation, we broadly categorize the non-verbal behaviors into bodily movements and facial expressions. Vocalic paralinguistic cues are excluded from the present analysis, as the study focuses on response scenarios involving solely non-verbal behaviors.
In particular, facial expressions encompass movements of the facial muscles, such as eyebrow raises, eye rolling, and changes in the corners of the mouth. The face, as a key component of the visual channel, serves as a powerful means of information transmission. Movements focus on bodily actions rather than facial ones, including body gestures, body postures, proxemics, haptics, and other actions.
We further introduce a silence category based on its unique informational characteristics. Although silence, which is the absence of any expression, action, or speech, inherently contains no information, humans paradoxically interpret it within conversations to convey meanings such as approval, displeasure, or agreement depending on the context~\cite{bruneau1973communicative,johannesen1974functions,jensen1973communicative}. Namely, silence serves to convey meaning despite its lack of explicit information, justifying its classification as a distinct category.

By categorizing non-verbal behaviors into these \textit{three distinct groups of silence, facial expressions, and movements}, this study offers a novel investigation into how effectively LLMs can interpret their implications when these behaviors serve as the sole form of response in a conversation.

\section{Detailed Problem Setting}\label{app:assump}
While the problem scope is introduced in the main text, this section offers a more detailed elaboration. In addition to the setting where the dialogue response consists solely of non-verbal behaviors, we establish two fundamental premises. First, we acknowledge that certain non-verbal behaviors may occur due to non-interactional factors (e.g., fatigue, distraction). To control for this, each question explicitly specifies that both participants are actively engaged in the conversation, constraining the non-verbal behaviors to occur solely as responses to speaker A's utterances. Namely, we incorporate the following elements into the question, eliminating ambiguity about intentionality: ``Assume that both parties faithfully engage in the conversation''.

Second, literal explanations of the conversational context are not inherently incorrect. However, we restrict the setting to ensure that the non-verbal responses contain latent meanings, in order to assess whether LLMs possess pragmatic competence. To make this setting clearly, we explicitly instruct the model to select the response that \textit{best interprets the context and underlying intent of the conversation}, as follows: ``Choose the option that most appropriately interprets the context and underlying intent of the conversation below.''

\section{Dataset Construction}\label{app:data_construction}
\paragraph{Source Data}
For rigorous evaluation, we construct a held-out evaluation set that shares no samples with any existing training data. Adopting the approaches of \citet{cui-etal-2020-mutual} and \citet{hu-etal-2023-fine}, we generate 50 self-collected situational prompts based on the empirically grounded observations of real-world interactions and GPT-4o outputs. These scenarios are designed to reflect pragmatically rich contents that commonly occur in daily lives. Each response in the dialogue is annotated with multiple behavioral categories, including silence, facial expressions, movements, and verbal behavior. By systematically varying only the response category, we aim to examine how different forms of non-verbal signals influence LLM interpretation within identical contexts.

\paragraph{Distractor Configuration}
To further ensure the difficulty of the benchmark, we construct multiple carefully designed distractor options for each example. The distractors are designed to capture typical pitfalls or failure patterns in pragmatic interpretation, including the following: 
\begin{itemize}[leftmargin=0pt]
\item \textit{Misinterpretations}, where the intended meaning is reversed or inaccurately inferred (e.g., interpreting an approving smile as disapproval)~\cite{sravanthi-etal-2024-pub}.
\item \textit{Literal explanations}, in which the understanding of the conversational context is correct but the response is described merely as a surface-level behavior (e.g., ``B smiled'' or ``B remained silent''), without attributing any pragmatic meaning, and in some cases, the non-verbal behavior is disregarded entirely~\cite{hu-etal-2023-fine}.
\item \textit{Lexically coherent but semantically divergent responses}, which share surface-level lexical similarity with the original context while lacking contextual relevance~\cite{shisen-etal-2024-large,zheng-etal-2021-grice}.
\item \textit{Abstract explanations}, which provide vague or generalized descriptions that lack contextual specificity and fail to address the core topic of the dialogue (e.g., stating only that ``A asks a question and B responds with a behavior'')~\cite{peters2025generalization}.
\end{itemize}

\paragraph{Quality Control}\label{app:quality_control}
We implement a rigorous quality control procedure on the generated dataset, which is conducted in two stages: automated assessment by LLMs followed by human cross-checking. In the LLM evaluation stage, the following criteria are employed: (1) Clarity: The dialogue should be clear, grammatical, and immediately understandable. (2) Correctness: The dialogue context and the designated answer must be properly aligned. (3) Difficulty: Among multiple plausible choices, the most appropriate response must be selected in accordance with the question’s intent. Based on these criteria, the LLMs evaluate quality scores on a 0–3 scale and provide a justification for each score: (1) 0 points: Poor dataset; both the dialogue and the options are entirely flawed. (2) 1 point: Major errors; the answer is not clearly correct. (3) 2 points: Minor errors, but the correct answer is clearly identifiable. (4) 3 points: Well-constructed dataset.

To enhance fairness, two different models (GPT-4o and GPT-4.1-mini) are employed for evaluation, and samples that consistently receive scores of 0 or 1 from both models are extracted. Subsequently, two human annotators review these samples, incorporate revisions informed by the LLM’s reasoning, and perform a cross-check to finalize the dataset.

Table~\ref{tb:stats} presents the statistics of the constructed dataset. We plan to release the dataset under the CC BY-SA 4.0 license, which permits redistribution and adaptation, provided that appropriate credit is given and derivative works are distributed under the same license.
\begin{table}[]\centering
\resizebox{.5\textwidth}{!}{%
\begin{tabular}{lccccc}
\toprule[1.5pt]
\textbf{Category} & \textbf{\# Samples} & \textbf{Avg. DL} & \textbf{Avg. OL} & \textbf{Avg. DL-Token} & \textbf{Avg. OL-Token} \\\midrule[1.5pt]
Silence & 50 & 68.26 & 70.79 & 14.3 & 13.14 \\
\begin{tabular}[c]{@{}l@{}}Facial\\ Expressions\end{tabular} & 50 & 68.22 & 84.92 & 14.24 & 15.2 \\
Movements & 50 & 92.68 & 74.8 & 17.22 & 13.74 \\
Verbal response & 50 & 96.2 & 80.5 & 19.12 & 14.35 \\ \bottomrule[1.5pt]
\end{tabular}%
}
\caption{Dataset statistics for a pragmatics-driven probing dataset. DL and OL denote dialogue length and option length, respectively.}
\label{tb:stats}
\end{table}

\end{document}